%% file: main.tex
\documentclass{article}
\usepackage[nonatbib,final]{nips_2017}
\usepackage{times}

\usepackage{amsmath,amssymb,amsthm,bm,multicol,graphicx}
\usepackage[colorlinks=true]{hyperref}
\usepackage{bibspacing}
\usepackage{multirow}
\usepackage{graphicx,color}
\usepackage{subfigure}
\usepackage[noend]{algpseudocode}
\usepackage{algorithmicx}
\usepackage{algorithm}
\usepackage{wrapfig}

\newcommand{\bbR}{\mathbb{R}}
\newcommand{\bmb}{{\bm b}}
\newcommand{\bmu}{{\bm u}}
\newcommand{\bmv}{{\bm v}}
\newcommand{\bmx}{{\bm x}}
\newcommand{\bmy}{{\bm y}}
\newcommand{\bmxi}{{\bm \xi}}
\newcommand{\bmzero}{{\bm 0}}
\newcommand{\bmg}{{\bm g}}
\newcommand{\bmeta}{{\bm \eta}}
\newcommand{\set}[1]{\{#1\}}

\title{Spectral Norm Regularization for Improving the Generalizability of Deep Learning}

\author{Yuichi Yoshida\\
  National Institute of Informatics\\
  \texttt{yyoshida@nii.ac.jp}
  \And
  Takeru Miyato\\
  Preferred Networks, Inc.\\
  \texttt{miyato@preferred.jp}
}

\begin{document}
\maketitle

\begin{abstract}
  We investigate the generalizability of deep learning based on the sensitivity to input perturbation.
  We hypothesize that the high sensitivity to the perturbation of data degrades the performance on it.
  To reduce the sensitivity to perturbation, we propose a simple and effective regularization method, referred to as spectral norm regularization, which penalizes the high spectral norm of weight matrices in neural networks.
  We provide supportive evidence for the abovementioned hypothesis by experimentally confirming that the models trained using spectral norm regularization exhibit better generalizability than other baseline methods.
\end{abstract}

\input{intro}

\input{related}

\input{method}
\input{experiments}
\input{conclusions}

\section*{Acknowledgement}
The authors thank Takuya Akiba and Seiya Tokui for helpful discussions.

\bibliographystyle{abbrvurl}
{
  \small
  \bibliography{main}
}

\appendix

\input{appendix}

\end{document}

%% file: intro.tex

\section{Introduction}

Deep learning has been successfully applied to several machine learning tasks including visual object classification~\cite{he2016identity,Krizhevsky:2012wl}, speech recognition~\cite{Hinton:2012is}, and natural language processing~\cite{Collobert:2008kg, jozefowicz2016exploring}.
A well-known method of training deep neural networks is stochastic gradient descent (SGD).
SGD can reach a local minimum with high probability over the selection of the starting point~\cite{Lee:2016vd}, and all local minima attain similar loss values~\cite{Choromanska:2015ui,Dauphin:2014tn,Kawaguchi:2016ub}.
However, the performance on test data, that is, \emph{generalizability}, can be significantly different among these local minima.
Despite the success of deep learning applications, we lack the understanding of generalizability, even though there has been progress ~\cite{Keskar:2017tz,Zhang:2017te}.

While understanding the generalizability of deep learning is an interesting topic, it is important from a practical point of view.
For example, suppose that we are training a deep neural network using SGD and want to parallelize the process using multiple GPUs or nodes to accelerate the training.
A well-known method for achieving this is synchronous SGD~\cite{das2016distributed, chen2016revisiting}, which requires a large minibatch to effectively exploit parallel computation on multiple GPUs or nodes.
However, it is reported that models trained through synchronous SGD with large minibatches exhibit poor generalization~\cite{Keskar:2017tz}, and a method is required to resolve this problem.

In this study, we consider the generalizability of deep learning from the perspective of sensitivity to input perturbation.
Intuitively, if a trained model is insensitive or sensitive to the perturbation of an input, then the model is confident or not confident about the output, respectively.
As the performance on test data is important, models that are insensitive to the perturbation of \emph{test} data are required.
Note that adversarial training~\cite{Szegedy:2014vw,Goodfellow:2015tl} is designed to achieve insensitivity to the perturbation of \emph{training} data, and it is not always effective for achieving insensitivity to the perturbation of test data.

To obtain insensitivity to the perturbation of test data, we propose a simple and effective regularization method,  referred to as spectral norm regularization.
As the name suggests, spectral norm regularization prevents the weight matrices used in neural networks  from having large spectral norms.
Through this, even though test data are not known in advance, a trained model is ensured to exhibit slight sensitivity to the perturbation of test data.

Using several real-world datasets, we experimentally confirm that models trained using spectral norm regularization exhibit better generalizability than  models trained using other baseline methods.
It is claimed in~\cite{Keskar:2017tz} that the maximum eigenvalue of the Hessian predicts the generalizability of a trained model.
However, we show that the insensitivity to the perturbation of test data is a more important factor for predicting generalizability, which further motivates the use of spectral norm regularization.
Finally, we show that spectral norm regularization effectively reduces the spectral norms of weight matrices.

The rest of this paper is organized as follows:
We review related works in Section~\ref{sec:related}.
In Section~\ref{sec:method}, we explain spectral norm regularization and compare it with other regularizing techniques .
Experimental results are provided in Section~\ref{sec:experiments}, and conclusions are stated in Section~\ref{sec:conclusions}.


%% file: related.tex

\section{Related Works}\label{sec:related}



A conventional method of understanding the generalizability of a trained model is the notion of the flatness/sharpness of a local minimum~\cite{Hochreiter:1997wf}.
A local minimum is (informally) referred to as \emph{flat} if its loss value does not increase significantly when it is perturbed; otherwise, it is referred to as \emph{sharp}.
In general, the high sensitivity of a training function at a sharp local minimizer negatively affects the generalizability of the trained model.
In~\cite{Hochreiter:1997wf}, this is explained in more detail through  the minimum description length  theory, which states that statistical models that require fewer bits to describe generalize better~\cite{Rissanen:1983fr}.

It is known that SGD with a large minibatch size leads to a model that does not generalize well~\cite{lecun2012efficient}.
In~\cite{Keskar:2017tz}, this problem is studied based on the flatness/sharpness of the obtained local minima.
They formulated a flat local minimum as a local minimum at which all eigenvalues of the Hessian are small (note that all eigenvalues are non-negative at a local minimum). Using a proxy measure, they experimentally showed that SGD with a smaller minibatch size tends to converge to a flatter minimum.

The notion of flat/sharp local minima considers the sensitivity of a loss function against the perturbation of model parameters.
However, it is natural to consider the sensitivity of a loss function against the perturbation of input data, as we discuss in this paper.
In~\cite{Szegedy:2014vw}, the perturbation to training data that increases the loss function the most is considered, and the resulting perturbed training data are referred to as \emph{adversarial examples}.
It is reported in~\cite{Goodfellow:2015tl} that training using adversarial examples improves test accuracy.

Recently,~\cite{Zhang:2017te} showed that the  classical notions of Rademacher complexity and the VC  dimension are not adequate for understanding the generalizability of deep neural networks. 

Note that the spectral norm of a matrix is equal to its largest singular value.
Singular values have attracted attention in the context of training recurrent neural networks (RNN).
In~\cite{Arjovsky:2016tb,Wisdom:2016vk}, it is shown that by restricting the weight matrices in RNN to be unitary or orthogonal, that is, matrices with all singular values equal to one, the problem of diminishing and exploding gradients can be prevented and better performance  can be obtained.

%% file: method.tex

\section{Spectral Norm Regularization}\label{sec:method}

In this section, we explain spectral norm regularization and how it reduces the sensitivity to test data perturbation.

\subsection{General idea}\label{subsec:intuition}
We consider feed-forward neural networks as a simple example to explain the intuition behind spectral norm regularization.
A feed-forward neural network can be represented as cascaded computations, $\bmx^\ell = f^\ell(W^\ell\bmx^{\ell-1} + \bmb^\ell)$ for $\ell = 1,\ldots,L$ for some $L$, where $\bmx^{\ell-1} \in \bbR^{n_{\ell-1}}$ is the input feature of the $\ell$-th layer, $f^\ell:\bbR^{n_\ell}  \to \bbR^{n_\ell}$ is a (non-linear) activation function, and $W^\ell \in \bbR^{n_{\ell} \times n_{\ell-1}}$ and $b^\ell \in \bbR^{n_\ell}$ are the layer-wise weight matrix and bias vector, respectively.
For a set of parameters, $\Theta = \set{W^\ell,\bmb^\ell}_{\ell=1}^L$, we denote by $f_\Theta:\bbR^{n_0} \to \bbR^{n_L}$ the function defined as $f_\Theta(\bmx^0) = \bmx^L$.
Given training data, $(\bmx_i,\bmy_i)_{i=1}^K$, where $\bmx_i \in \bbR^{n_0}$ and $\bmy_i \in \bbR^{n_L}$, the loss function is defined as $\frac{1}{K}\sum_{i=1}^K L(f_\Theta(\bmx_i),\bmy_i)$, where $L$ is frequently selected to be cross entropy and the squared $\ell_2$-distance for classification and regression tasks, respectively.
The model parameter to be learned is $\Theta$.

Let us consider how we can obtain a model that is insensitive to the perturbation of the input.
Our goal is to obtain a model, $\Theta$, such that the $\ell_2$-norm of $f(\bmx + \bmxi) - f(\bmx)$ is small, where $\bmx \in \bbR^{n_0}$ is an arbitrary vector and $\bmxi \in \bbR^{n_0}$ is a perturbation vector with a small $\ell_2$-norm.
A key observation is that most practically used neural networks exhibit nonlinearity only because they use piecewise linear functions, such as ReLU , maxout~\cite{Goodfellow:2013tf}, and maxpooling~\cite{Ranzato:2007eb}, as activation functions.
In such a case, function $f_\Theta$ is a piecewise linear function.
Hence, if we consider a small neighborhood of $\bmx$, we can regard $f_\Theta$ as a linear function. In other words, we can represent it by an affine map, $\bmx \mapsto W_{\Theta,\bmx}\bmx + \bmb_{\Theta,\bmx}$, using a matrix, $W_{\Theta,\bmx}  \in \bbR^{n_0\times n_L}$, and a vector, $\bmb_{\Theta,\bmx} \in \bbR^{n_L}$, which depend on $\Theta$ and $\bmx$.
Then, for a small perturbation, $\bmxi \in \bbR^{n_0}$, we have
\begin{align*}
  \frac{\|f_\Theta(\bmx + \bmxi) - f(\bmx)\|_2}{\|\bmxi\|_2}
  = \frac{\|(W_{\Theta,\bmx}(\bmx + \bmxi) + \bmb_{\Theta,\bmx}) - (W_{\Theta,\bmx}\bmx + \bmb_{\Theta,\bmx}) \|_2 }{\|\bmxi\|_2}
  = \frac{\|W_{\Theta,\bmx}\bmxi\|_2}{\|\bmxi\|_2} \leq \sigma(W_{\Theta,\bmx}),
\end{align*}
where $\sigma(W_{\Theta,\bmxi})$ is the spectral norm of $W_{\Theta,\bmxi}$.
The \emph{spectral norm} of a matrix $A \in \bbR^{m \times n}$ is defined as
\[
  \sigma(A) = \max_{\bmxi \in \bbR^n,\bmxi \neq \bmzero}\frac{\|A \bmxi\|_2}{\|\bmxi\|_2},
\]
which corresponds to the largest singular value of $A$.
Hence, the function $f_\Theta$ is insensitive to the perturbation of $\bmx$ if the spectral norm of $W_{\Theta,\bmx}$ is small.

The abovementioned argument suggests that model parameter $\Theta$ should be trained so that the spectral norm of $W_{\Theta,\bmx}$ is small for any $\bmx$.
To further investigate the property of $W_{\Theta,\bmx}$, let us assume that each activation function, $f^\ell$, is an element-wise ReLU (the argument can be easily generalized to other piecewise linear functions).
Note that, for a given vector, $\bmx$, $f^\ell$ acts as a diagonal matrix, $D^\ell_{\Theta,\bmx} \in \bbR^{n_\ell \times n_{\ell}}$, where an element in the diagonal is equal to one if the corresponding element in $\bmx^{\ell-1}$ is positive; otherwise, it is equal to zero.
Then, we can rewrite $W_{\Theta,\bmx}$ as $W_{\Theta,\bmx} = D^{L}_{\Theta,\bmx}W^{L}D^{L-1}_{\Theta,\bmx} W^{L-1} \cdots D^1_{\Theta,\bmx} W^1$.
Note that $\sigma(D^\ell_{\Theta,\bmx}) \leq 1$ for every $\ell \in \set{1,\ldots,L}$.
Hence, we have
\[
  \sigma(W_{\Theta,\bmx})
  \leq
  \sigma(D^{L}_{\Theta,\bmx})\sigma(W^{L})\sigma(D^{L-1}_{\Theta,\bmx}) \sigma(W^{L-1}) \cdots \sigma(D^{1}_{\Theta,\bmx}) \sigma(W^1)
  \leq
  \prod_{\ell=1}^L \sigma(W^\ell).
\]
It follows that, to bound the spectral norm of $W_{\Theta,\bmx}$, it suffices to bound the spectral norm of $W^\ell$ for each $\ell \in \set{1,\ldots,L}$.
This motivates us to consider spectral norm regularization, which is described in the next section.

\subsection{Details of spectral norm regularization}\label{subsec:proposed}

In this subsection, we explain spectral norm regularization.
The notations are the same as those used in Section~\ref{subsec:intuition}.
To bound the spectral norm of each weight matrix, $W^\ell$, we consider the following empirical risk minimization problem:
\begin{align}
  \mathop{\text{minimize}}_{\Theta} \frac{1}{K}\sum_{i=1}^K L(f_\Theta(\bmx_i),\bmy_i) + \frac{\lambda}{2} \sum_{\ell=1}^L \sigma(W^\ell)^2,
  \label{eq:spectral}
\end{align}
where $\lambda \in \bbR_+$ is a regularization factor.
We refer to the second term as the \emph{spectral norm regularizer}. It decreases the spectral norms of the weight matrices.

When performing SGD, we need to calculate the gradient of the spectral norm regularizer.
To this end, let us consider the gradient of $\sigma(W^\ell)^2_2/2$ for a particular $\ell \in \set{1,2,\ldots,L}$.
Let $\sigma_1 = \sigma(W^\ell)$ and $\sigma_2$ be the first and second singular values, respectively.
If $\sigma_1 > \sigma_2$, then the gradient of $\sigma(W^\ell)^2/2$ is $\sigma_1 \bmu_1\bmv_1^\top$, where $\bmu_1$ and $\bmv_1$ are the first left and right singular vectors, respectively.
If $\sigma_1 = \sigma_2$, then $\sigma(W^\ell)^2_2$ is not differentiable.
However, for practical purposes, we can assume that this case never occurs because numerical errors prevent $\sigma_1 $ and $\sigma_2$ from being exactly equal.

As it is computationally expensive to compute $\sigma_1$, $\bmu_1$, and $\bmv_1$, we approximate them using the power iteration method.
Starting with a randomly initialized $\bmv \in \bbR^{n_{\ell-1}}$, we iteratively perform the following procedure a sufficient number of times: $\bmu \leftarrow W^\ell \bmv$ and $\bmv \leftarrow (W^\ell)^\top \bmu$, and $\sigma \leftarrow \|\bmu\|_2/\|\bmv\|_2$.
Then, $\sigma$, $\bmu$, and $\bmv$ converge to $\sigma_1$, $\bmu_1$, and $\bmv_1$, respectively (if $\sigma_1 > \sigma_2$).
To approximate $\sigma_1$, $\bmu_1$, and $\bmv_1$ in the next iteration of SGD, we can reuse $\bmv$ as the initial vector.
in our experiments, which are explained in Section~\ref{sec:experiments}, we performed only one iteration because it was adequate for obtaining a sufficiently good approximation.
A pseudocode is provided in Algorithm~\ref{alg:spectral}.

\begin{algorithm}[t!]
  \caption{SGD with spectral norm regularization}\label{alg:spectral}
  \begin{algorithmic}[1]
    \For{$\ell=1$ to $L$}
      \State $\bmv^\ell \leftarrow$ a random Gaussian vector.
    \EndFor
    \For{each iteration of SGD}
      \State Consider a minibatch, $\set{(\bmx_{i_1},y_{i_1}),\ldots,(\bmx_{i_k},y_{i_k})}$, from training data.
      \State Compute the gradient of $\frac{1}{k}\sum_{i=1}^k L(f_{\Theta}(\bmx_{i_j}),y_{i_j})$ with respect to $\Theta$.
      \For{$\ell=1$ to $L$}
        \For{a sufficient number of times} \Comment{One iteration was adequate in our experiments}
          \State $\bmu^\ell \leftarrow W^\ell \bmv^\ell$, $\bmv^\ell \leftarrow (W^\ell)^\top \bmu^\ell$, $\sigma^\ell \leftarrow \|\bmu^\ell\|/\|\bmv^\ell\|$
          \State Add $\lambda \sigma^\ell \bmu^\ell (\bmv^\ell)^\top$ to the gradient of $W^\ell$.
        \EndFor
      \EndFor
      \State Update $\Theta$ using the gradient.
    \EndFor
  \end{algorithmic}
\end{algorithm}

\paragraph{Convolutions.}
We describe how to handle convolutions because they are used widely in recent applications of  deep neural networks.
Consider a convolutional layer with $a$ input channels, $b$ output channels, and a $k_w \times k_h$-sized kernel.
This implies that the convolution has $abk_wk_h$ parameters.
Note that a value in an output channel is determined using $ak_wk_h$ values in the input channels.
Hence, we align the parameters as a matrix of size $b \times ak_wk_h$ and apply the abovementioned power iteration method to the matrix to calculate its spectral norm and gradient.



\subsection{Comparison with other regularization methods}
We now compare the spectral norm regularization with other regularization techniques.
\paragraph{Weight decay.}

\emph{Weight decay}, or the \emph{Frobenius norm regularization}, is a well-known regularization technique for deep learning.
It considers the following problem:
\begin{align}
  \mathop{\text{minimize}}_{\Theta} \frac{1}{K}\sum_{i=1}^K L(f_\Theta(\bmx_i),\bmy_i) + \frac{\lambda}{2}\sum_{\ell=1}^L\|W^\ell\|_F^2,
  \label{eq:weight-decay}
\end{align}
where $\lambda \in \bbR_+$ is a regularization factor.
We note that $\|W^\ell\|_F^2 = \sum_{i=1}^{\min\set{n_{\ell-1},n_\ell}}\sigma_i(W^\ell)^2$, where $\sigma_i(W^\ell)$ is the $i$-th singular value of $W^\ell$.
Hence, the Frobenius norm regularization reduces the sum of squared singular values.
Even though it will be effective to train models that are  insensitive to input perturbation, a trained model may lose important information about the input because each trained weight matrix, $W^\ell$, acts as an operator that shrinks in all directions.
In contrast, spectral norm regularization focuses only on the first singular value, and each trained weight matrix, $W^\ell$, does not shrink significantly in the directions orthogonal to the first right singular vector.

\paragraph{Adversarial training.}

Adversarial training~\cite{Goodfellow:2015tl} considers the following problem:
\begin{align}
  \mathop{\text{minimize}}_{\Theta} \alpha \cdot \frac{1}{K}\sum_{i=1}^K L(f_\Theta(\bmx_i),\bmy_i) + (1-\alpha) \cdot \frac{1}{K}\sum_{i=1}^K L(f_\Theta(\bmx_i + \bmeta_i), \bmy_i)\label{eq:adversarial},
\end{align}
where
\[
  \bmeta_i = \epsilon \cdot \frac{\bmg_i}{\|\bmg_i\|_2} \quad \text{and} \quad \bmg_i = \nabla_{\bmx} L(f_{\Theta}(\bmx),\bmy_i)|_{\bmx = \bmx_i},
\]
and $\alpha$ and $\epsilon \in \bbR_+$ are hyperparameters.
It considers the perturbation toward the direction that increases the loss function the most.
Hence, a trained model is insensitive to the adversarial perturbation of training data.
In contrast, spectral norm regularization automatically trains a model that is insensitive to the perturbation of training data and test data.

\paragraph{Jacobian regularization.}
Another method of reducing the sensitivity of a model against input perturbation is penalizing the $\ell_2$-norm of the derivative of the output with respect to the input.
Let us denote the Jacobian matrix of $\bmy$ with respect to $\bmx$ by $\partial \bmy / \partial \bmx$.
The \emph{Jacobian regularization} considers the following regularization term:
\begin{align*}
  \frac{1}{K}\sum_{i=1}^K \left\|\frac{\partial \bmy}{\partial \bmx}\Big|_{\bmx=\bmx_i}\right\|_F^2.
\end{align*}
The Jacobian regularization promotes the smoothness of a model against input perturbation.
However, this regularization is impractical because calculating the derivative of a Jacobian with respect to parameters is computationally expensive.
To resolve the issue, Gu~\emph{et~al.}~\cite{gu2014towards} proposed an alternative method that regularizes layer-wise Jacobians:
\[
  \frac{1}{K}\sum_{i=1}^K \sum_{\ell=1}^{L} \left\|\frac{\partial \bmx^{\ell}}{\partial \bmx^{\ell-1}}\Big|_{\bmx^{\ell-1}=\bmx^{\ell-1}_i}\right\|_F^2,
\]
where $\bmx^\ell_i$ is the input to the $\ell$-th layer calculated using $\bmx_i$.
Note that, if we neglect the effect of activation functions between layers, this regularization coincides with weight decay.
Hence, we exclude this regularization from our experiments.

%% file: experiments.tex
\section{Experiments}\label{sec:experiments}

In this section, we experimentally demonstrate the effectiveness of spectral norm regularization on classification tasks over other regularization techniques, and confirm that the insensitivity to test data perturbation is an important factor for generalization.

All the training methods discussed here are based on stochastic gradient descent (SGD).
We consider two regimes for the choice of the mini-batch size.
In the small-batch regime, we set the mini-batch size to $B=64$, and in the large-batch regime, we set it to $B=4096$.
In our experiments, we compared the following four problems:
\begin{itemize}
\itemsep=0pt
\item Vanilla problem (\textsf{vanilla}):
  As a vanilla problem, we considered empirical risk minimization without any regularization, that is, $\mathop{\text{minimize}}_{\Theta} \frac{1}{K}\sum_{i=1}^K L(f_\Theta(\bmx_i),\bmy_i)$, where $L$ is the cross entropy.
\item Weight decay (\textsf{decay}): We considered the problem~\eqref{eq:weight-decay}, where $L$ is the cross entropy.
We selected the regularization factor $\lambda=10^{-4}$.
\item Adversarial training (\textsf{adversarial}): We considered the problem~\eqref{eq:adversarial}, where $L$ is the cross entropy.
We selected $\alpha=0.5$  and $\epsilon=1$, as suggested in~\cite{Goodfellow:2015tl}.
\item Spectral norm regularization (\textsf{spectral}): We considered the problem~\eqref{eq:spectral}, where $L$ is the cross entropy.
We selected the regularization factor $\lambda =0.01$.
\end{itemize}
We trained models using Nesterov's accelerated gradient descent~\cite{Bengio:2013fn} with momentum 0.9.
We decreased the learning rate by a factor of $1/10$ when the half and the three quarters of the training process have passed.
We optimized the hyper-parameters through a grid search and selected those that showed a reasonably good performance for every choice of neural network and mini-batch size.

In our experiments, we used the following four settings on the model and dataset.
\begin{itemize}
\itemsep=0pt
\item The VGG network (VGGNet for short)~\cite{Simonyan:2014ws} on the CIFAR-10 dataset~\cite{Krizhevsky:2009tr}.
\item The network in network (NIN) model ~\cite{Lin:2014wb} on the CIFAR-100 dataset~\cite{Krizhevsky:2009tr}.
\item The densely connected convolutional network (DenseNet) ~\cite{Huang:2016wa} having a depth of 40 on the CIFAR-100 dataset.
\item DenseNet having a depth of 22 on the STL-10 dataset~\cite{Coates:2011wo}.
  The depth was decreased because of the memory consumption issue.
\end{itemize}
We preprocessed all the datasets using global contrast normalization.
We further applied data augmentation with cropping and horizontal flip on STL-10 because the number of training data samples is only 5,000, which is small considering the large mini-batch size of 4096.
The learning rate of SGD was initialized to $0.1$ in the small-batch regime and $1.0$ in the large-batch regime for the NIN and DenseNet models on the CIFAR-100 dataset, and was initialized to $0.01$ in the small-batch regime and $0.1$ in the large-batch regime for the VGGNet and DenseNet models on the CIFAR-10 dataset.





\subsection{Accuracy and Generalization Gap}\label{subsec:accuracy}

\begin{table}
  \centering
  \caption{Test accuracy and generalization gap (best values in bold).}\label{tab:accuracy}
  {
  \small
  \tabcolsep=1mm
  \begin{tabular}{|c|r||rrrr||r|rrrr|}
    \hline
    & & \multicolumn{4}{c||}{Test accuracy} & \multicolumn{5}{c|}{Generalization gap} \\
    Model & $B$ & \textsf{vanilla} & \textsf{decay} & \textsf{adver.} & \textsf{spectral} & \multicolumn{1}{c}{$\alpha$} & \textsf{vanilla} & \textsf{decay} & \textsf{adver.} & \textsf{spectral}  \\ \hline \hline
    \multirow{2}{*}{VGGNet} & 64 & 0.898 & 0.897 & 0.884 & \textbf{0.904} & 0.88 & 0.079 & 0.074 & 0.109 & \textbf{0.068} \\
    & 4096 & 0.858 & 0.863 & 0.870 & \textbf{0.885} & 0.85 & 0.092 & 0.064 & 0.064 & \textbf{0.045} \\ \hline
    \multirow{2}{*}{NIN} & 64 & 0.626 & \textbf{0.672} & 0.627 & 0.669 & 0.62 & 0.231 & 0.120 & 0.253 & \textbf{0.090} \\
    & 4096 & 0.597 & 0.618 & 0.607 & \textbf{0.640} & 0.59 & 0.205 & 0.119 & 0.196 & \textbf{0.090} \\ \hline
    DenseNet & 64 & 0.675 & \textbf{0.718} & 0.675 & 0.709 & 0.67 & 0.317 & 0.080 & 0.299 & \textbf{0.095}\\
    (CIFAR100) & 4096 & 0.639 & 0.671 & 0.649 & \textbf{0.697} & 0.63 & 0.235 & 0.111 & 0.110 & \textbf{0.051}  \\ \hline
    DenseNet  & 64 & 0.724 & 0.723 & 0.707 & \textbf{0.735} & 0.70 & \textbf{0.063} & 0.073 & 0.069 & 0.068 \\
    (STL-10) & 4096 & 0.686 & 0.689 & 0.676 & \textbf{0.697} & 0.67 & 0.096 & 0.057 & \textbf{0.015} & 0.042 \\
    \hline
  \end{tabular}
  }
\end{table}


\begin{figure*}[t!]
  \subfigure[$B=64$]{
    \includegraphics[width=.48\hsize]{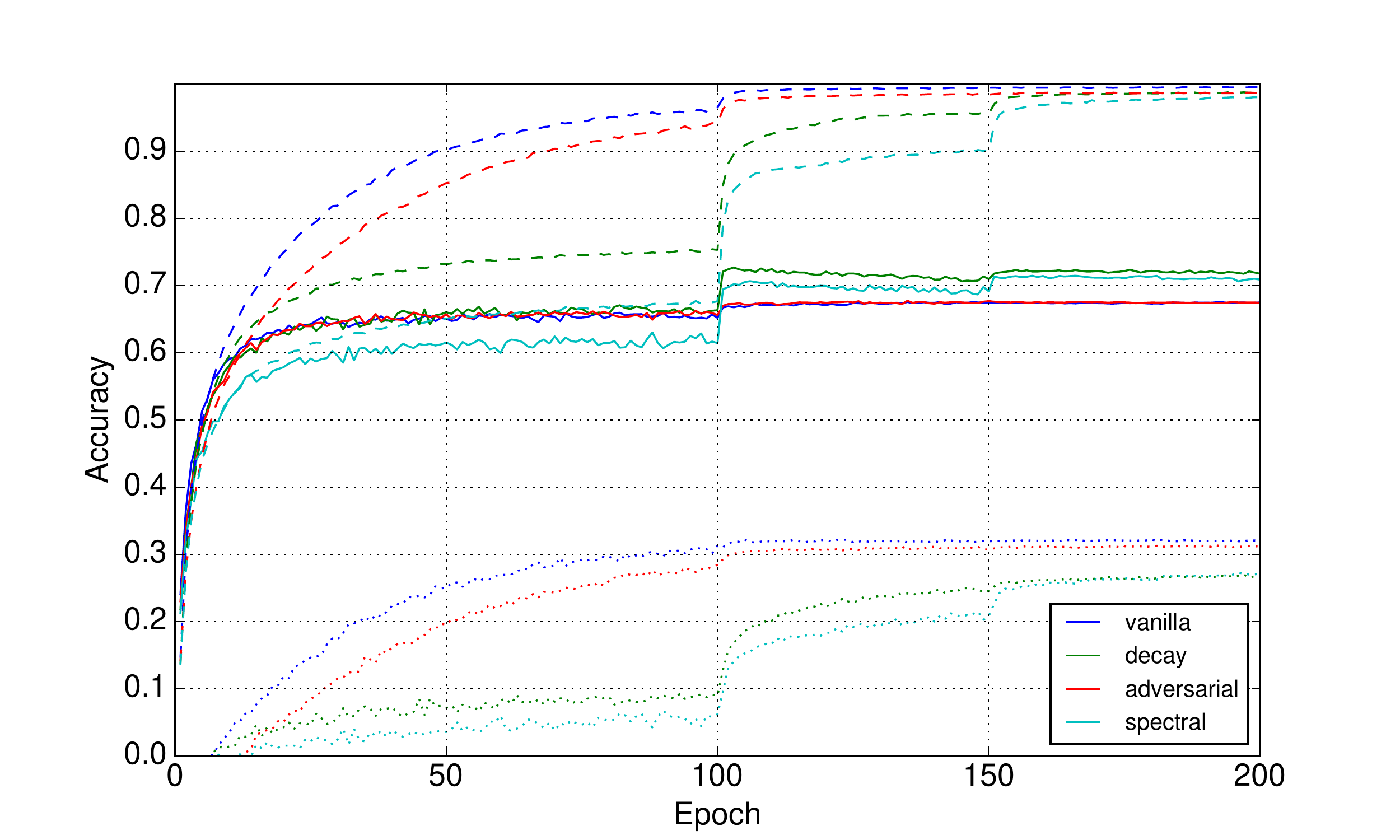}
  }
  \subfigure[$B=4096$]{
    \includegraphics[width=.48\hsize]{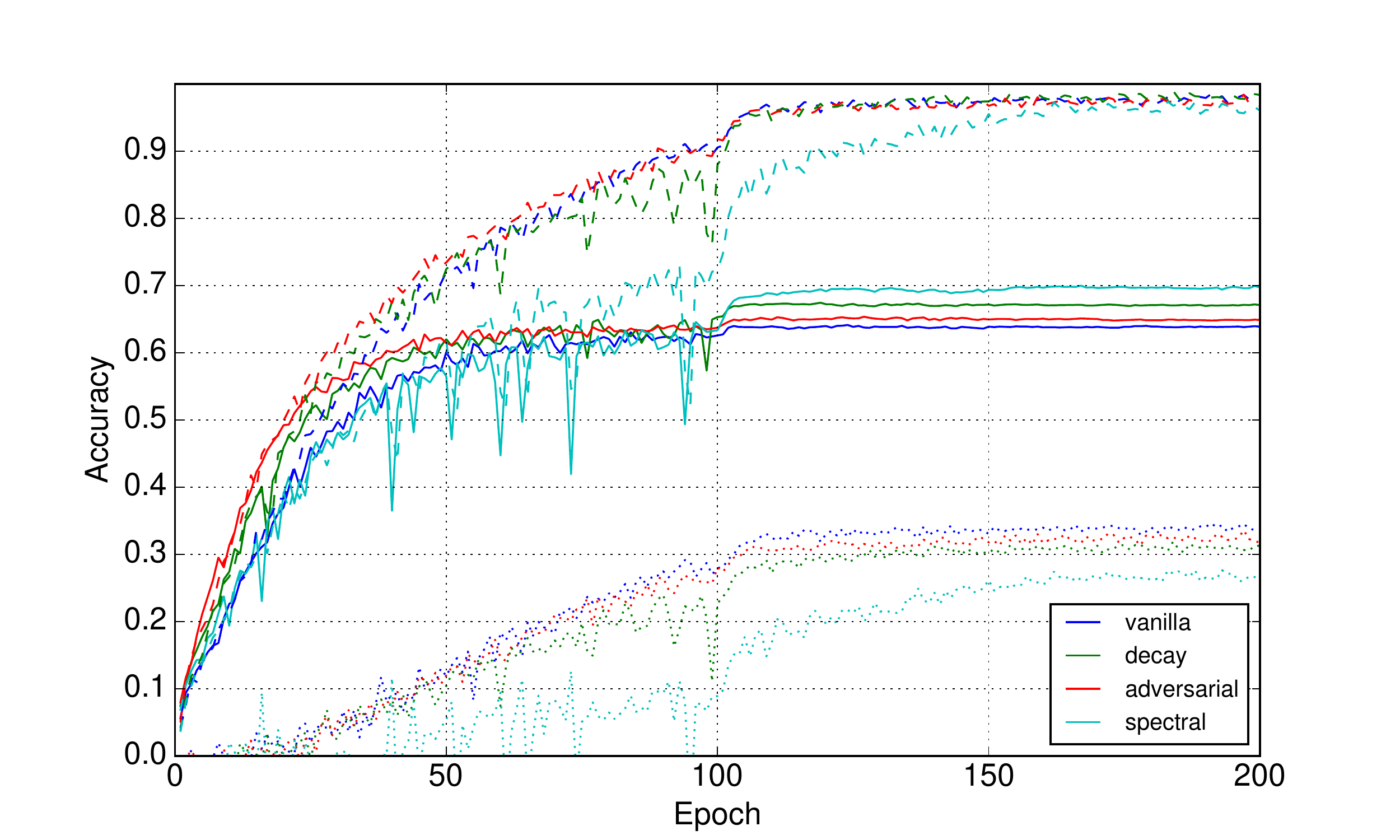}
  }
  \caption{Accuracy of the DenseNet model on the CIFAR-100 dataset. The solid, dashed, and dotted lines indicate the test accuracy, training accuracy, and generalization gap, respectively.}\label{fig:accuracy-DenseNet}
\end{figure*}

First, we look at the test accuracy obtained by each method, which is summarized in the left columns of Table~\ref{tab:accuracy}.
In the small-batch regime, \textsf{decay} and \textsf{spectral} show better test accuracies than the other two methods.
In the large-batch regime, \textsf{spectral} clearly achieves the best test accuracy for every model.
Although the test accuracy decreases as the mini-batch size increases, as reported in~\cite{Keskar:2017tz}, the decrease in the test accuracy of \textsf{spectral} is less significant than those of the other three methods.

Next, we look at the generalization gap.
We define the generalization gap at a threshold $\alpha$ as the minimum difference between the training and test accuracies when the test accuracy exceeds $\alpha$.
The generalization gap of each method is summarized in the right columns of Table~\ref{tab:accuracy}.
For each setting, we selected the threshold $\alpha$ so that every method achieves a test accuracy that (slightly) exceeds this threshold.
For each of the settings, except for the DenseNet model on the STL-10 dataset, \textsf{spectral} clearly shows the smallest generalization gap followed by \textsf{decay}, which validates the effectiveness of \textsf{spectral}.

Figure~\ref{fig:accuracy-DenseNet} shows the training curve of the DenseNet model on the CIFAR-100 dataset.
The results for other settings are given in Appendix~\ref{apx:accuracy}.
As we can observe, in both the small-batch and large-batch regimes, \textsf{spectral} shows the smallest generalization gaps.
The generalization gap of \textsf{decay} increases as the mini-batch size increases, whereas that of \textsf{spectral} does not increase significantly.
Investigating the reason behind this phenomena is an interesting future work.

The choice of the mini-batch size and the training method is not important to obtain a model with a good training accuracy; all of them exceeds 95\%.
However, obtaining a model with a good test accuracy, or a small generalization gap, is important.
In the subsequent sections, we investigate which property of a trained model determines its generalization gap.

To summarize, \textsf{spectral} consistently achieves the small generalization gap and shows the best test accuracy, especially in the large-batch regime.


\subsection{Sensitivity to the perturbation of the input}



\begin{figure}
  \centering
  \subfigure[Train]{
    \label{fig:grad-transition-train}
    \includegraphics[width=.475\hsize]{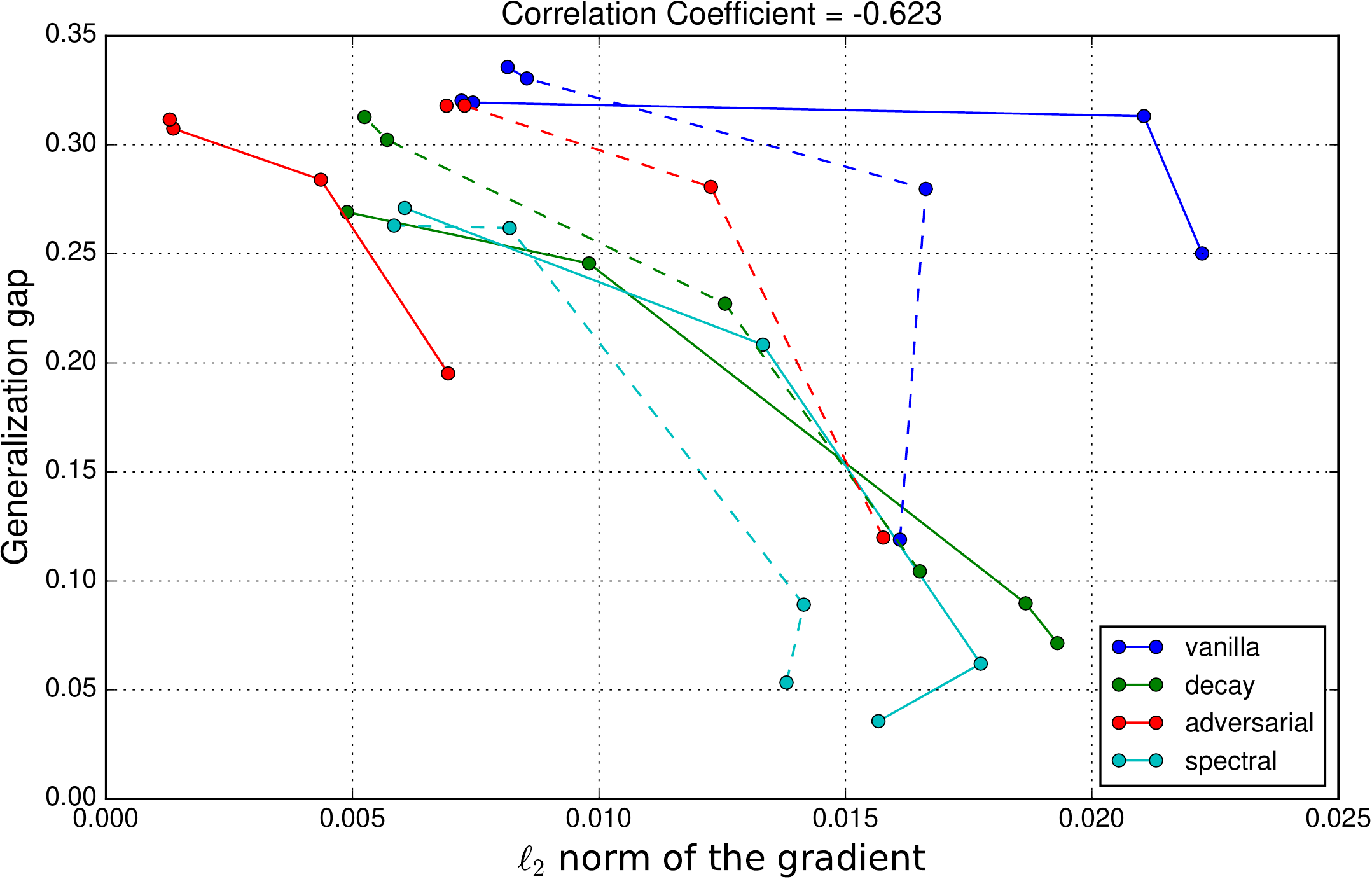}
  }
  \subfigure[Test]{
    \label{fig:grad-transition-test}
    \includegraphics[width=.475\hsize]{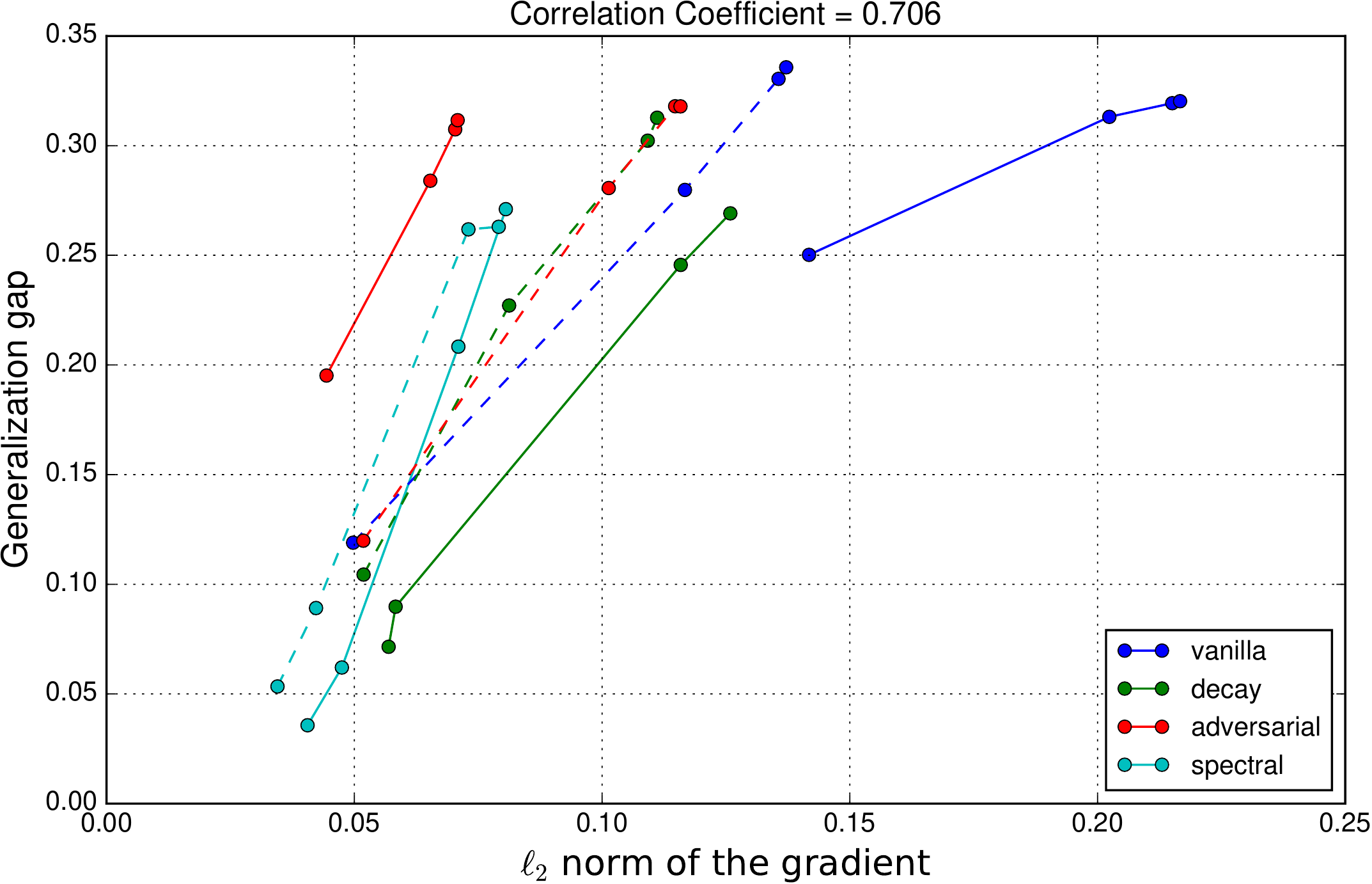}
  }
  \caption{Relation between the generalization gap and the $\ell_2$-norm of the gradient of the DenseNet model on the CIFAR-100 dataset. The solid and dashed lines indicate the results for the small-batch and large-batch regimes, respectively.}\label{fig:grad-transition}
\end{figure}

To understand the relation between the sensitivity to the input perturbation and the generalization gap of the trained model, we look at the gradient of the loss function, defined with the training data or the test data, with respect to the input.
Figure~\ref{fig:grad-transition} shows the transition of the $\ell_2$-norm when training the DenseNet model on the CIFAR-100 dataset.
The results for other settings are given in Appendix~\ref{apx:sensitivity}.

We can observe in Figure~\ref{fig:grad-transition-test} that the $\ell_2$-norm of the gradient with respect to the test data is well  correlated with the generalization gap.
In particular, the $\ell_2$-norm of the gradient gradually increases as training proceeds, which matches the behavior of the generalization gap, which also increases as training proceeds.

On the contrary, as shown in Figure~\ref{fig:grad-transition-train}, although the $\ell_2$-norm of the gradient gradually decreases as training proceeds, the generalization gap actually increases.
Hence, the $\ell_2$-norm of the gradient with respect to the training data does not predict the generalization gap well.

These results motivate us to reduce the $\ell_2$-norm of the gradient with respect to the test data to obtain a smaller generalization gap.
As we do not know the test data a priori, the spectral norm regularization method is reasonable to achieve this goal because it reduces the gradient at any point.

The superiority of \textsf{spectral} to \textsf{decay} can be elucidated from this experiment.
In order to achieve a better test accuracy, we must make the training accuracy high and the generalization gap small.
To achieve the latter, we have to penalize the $\ell_2$-norm of the gradient with respect to the test data.
To this end, \textsf{decay} suppresses all the weights, which decreases model complexity, and hence, we cannot fit the model to the training data well.
On the other hand, \textsf{spectral} only suppresses the spectral norm, and hence, we can achieve a greater model complexity than \textsf{decay} and fit the model to the training data better.



\subsection{Maximum eigenvalue of the Hessian with respect to the model parameters}

\begin{figure}
  \centering
  \begin{minipage}{.45\hsize}
  \subfigure[Train]{
    \includegraphics[width=0.46\hsize]{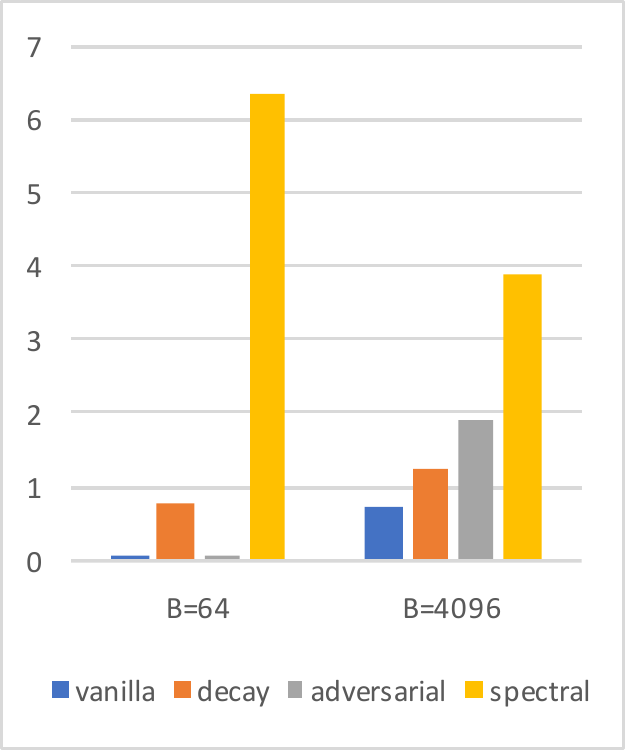}
  }
  \subfigure[Test]{
    \includegraphics[width=0.46\hsize]{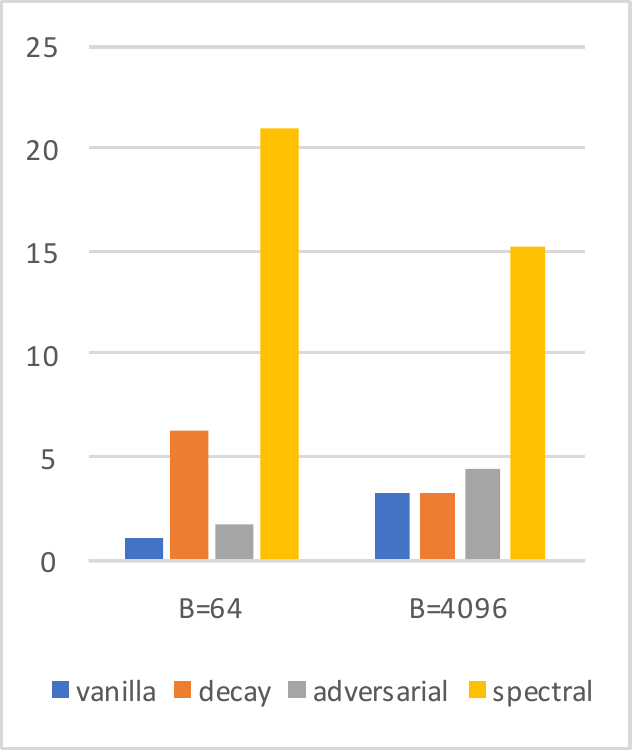}
  }
  \caption{Maximum eigenvalue of the Hessian of the DenseNet on CIFAR100}\label{fig:hessian}
  \end{minipage}
  \quad
  \begin{minipage}{.45\hsize}
  \includegraphics[width=\hsize]{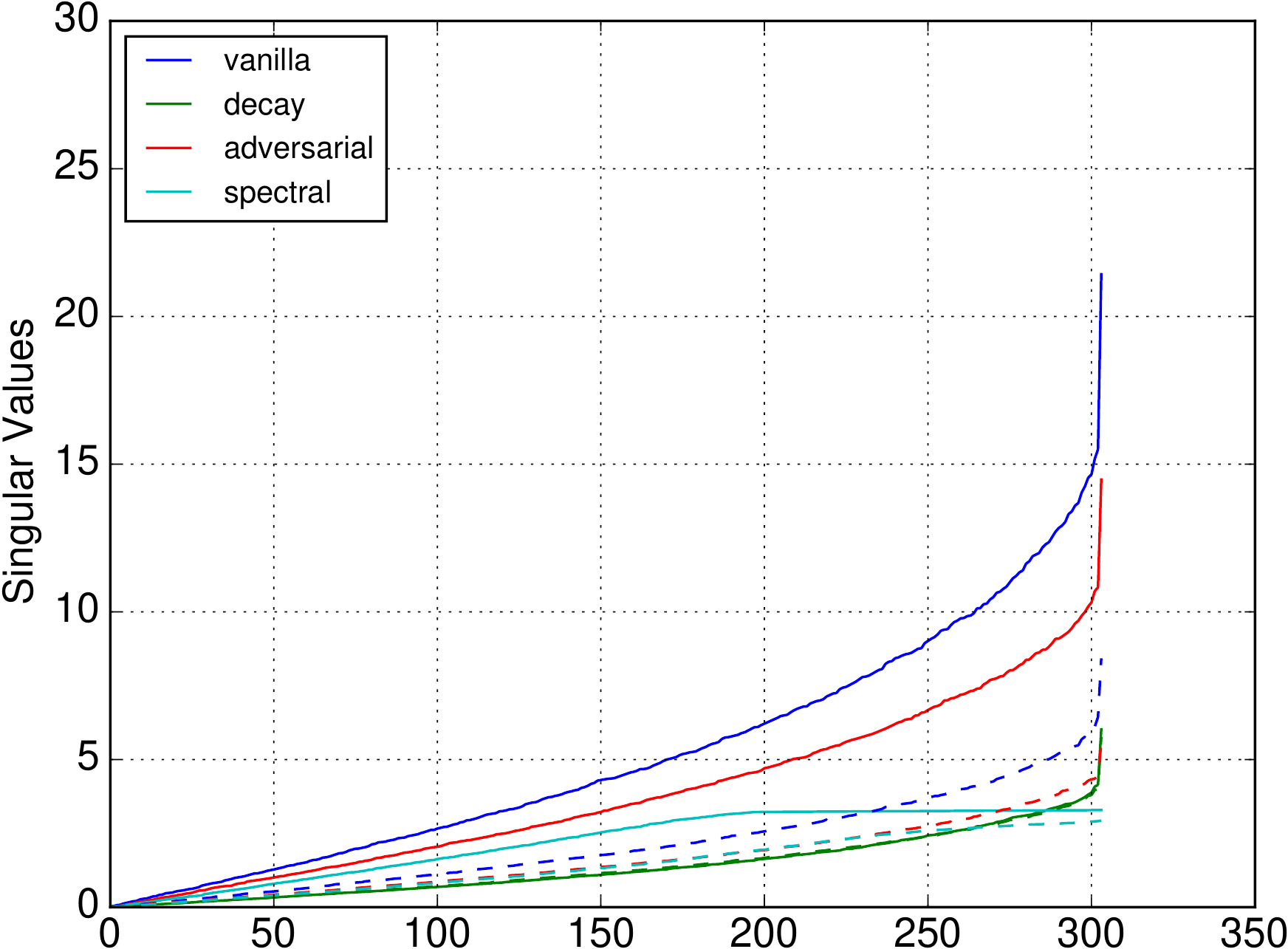}
  \caption{Singular values of weight matrices in the DenseNet on CIFAR100. Solid and dashed lines indicate the results for the small-batch and large-batch regimes, respectively.}\label{fig:spectrum}
  \end{minipage}
\end{figure}

In~\cite{Keskar:2017tz}, it is claimed that the maximum eigenvalue of the Hessian of the loss function defined with the training data predicts the generalization gap well.
To confirm this claim, we computed the maximum eigenvalue of the DenseNet model trained on the CIFAR-100 dataset, shown in Figure~\ref{fig:hessian}.
As it is computationally expensive to compute the Hessian, we approximated its maximum eigenvalue by using the power iteration method because we can calculate the Hessian-vector product without explicitly calculating the Hessian~\cite{Martens:2010vo}.
We also computed the maximum eigenvalue of the Hessian of the loss function defined with the test data.

We can observe that, for \textsf{vanilla}, larger eigenvalues (in both the training and test data) are obtained if the mini-batch size is increased, which confirms the claim of~\cite{Keskar:2017tz}.
However, the models trained with regularizations tend to have larger eigenvalues, although they have better generalizability.
In particular, the models trained by \textsf{spectral} have the largest eigenvalues, although they have the best generalizability as we have seen in Section~\ref{subsec:accuracy}.

This phenomena can be understood as follows:
If we train a model without regularization, then a small perturbation does not significantly affect the Frobenius and spectral norms of weight matrices because they are already large.
However, if we train a model with regularization, then because of these small norms a small perturbation may significantly affect those norms, which may cause a significant change in the output.

To summarize, this experiment indicates that the maximum eigenvalue of the Hessian of the loss function is not a suggestive measure for predicting generalizability.

\subsection{Singular values of weight matrices}

Finally, we look at the singular values of weight matrices in the models trained by each method.
Figures~\ref{fig:spectrum} shows the singular values of a weight matrix taken from the DenseNet model on the CIFAR-100 dataset.
The matrix is selected arbitrarily because all matrices showed similar spectra.

We can observe that the spectrum of \textsf{vanilla} is highly skewed, and \textsf{adversarial} and \textsf{decay} shrink the spectrum while maintaining the skewed shape.
In contrast, the spectrum of \textsf{spectral} is flat.
This behavior is as expected because the spectral norm regularization tries to reduce the largest singular value.
Because the maximum singular value obtained by \textsf{spectral} is low, we obtain less sensitivity to the perturbation of the input.


%% file: conclusions.tex

\section{Conclusions}\label{sec:conclusions}
In this work, we hypothesized that a high sensitivity to the perturbation of the input data degrades the performance of the data.
In order to reduce the sensitivity to the perturbation of the test data, we proposed the spectral norm regularization method, and confirmed that it exhibits a better generalizability than other baseline methods through experiments.
Experimental comparison with other methods indicated that the insensitivity to the perturbation of the test data plays a crucial role in determining the generalizability.

There are several interesting future directions to pursue.
It is known that weight decay can be seen as a regularization in MAP estimation derived from a Gaussian prior to the model parameters.
Is it possible to understand spectral norm regularization as a derivation of a prior?
We also need a theoretical understanding of the effect of spectral norm regularization on generalization.
It is known that, in some ideal cases, weight decay improves generalizability by preventing neural networks from fitting noises~\cite{Krogh:1991uo}.
Can we extend this argument to spectral norm regularization?


%% file: appendix.tex

\section{Accuracy}\label{apx:accuracy}


\begin{figure*}[t!]
  \subfigure[VGGNet ($B=64$)]{
    \includegraphics[width=.48\hsize]{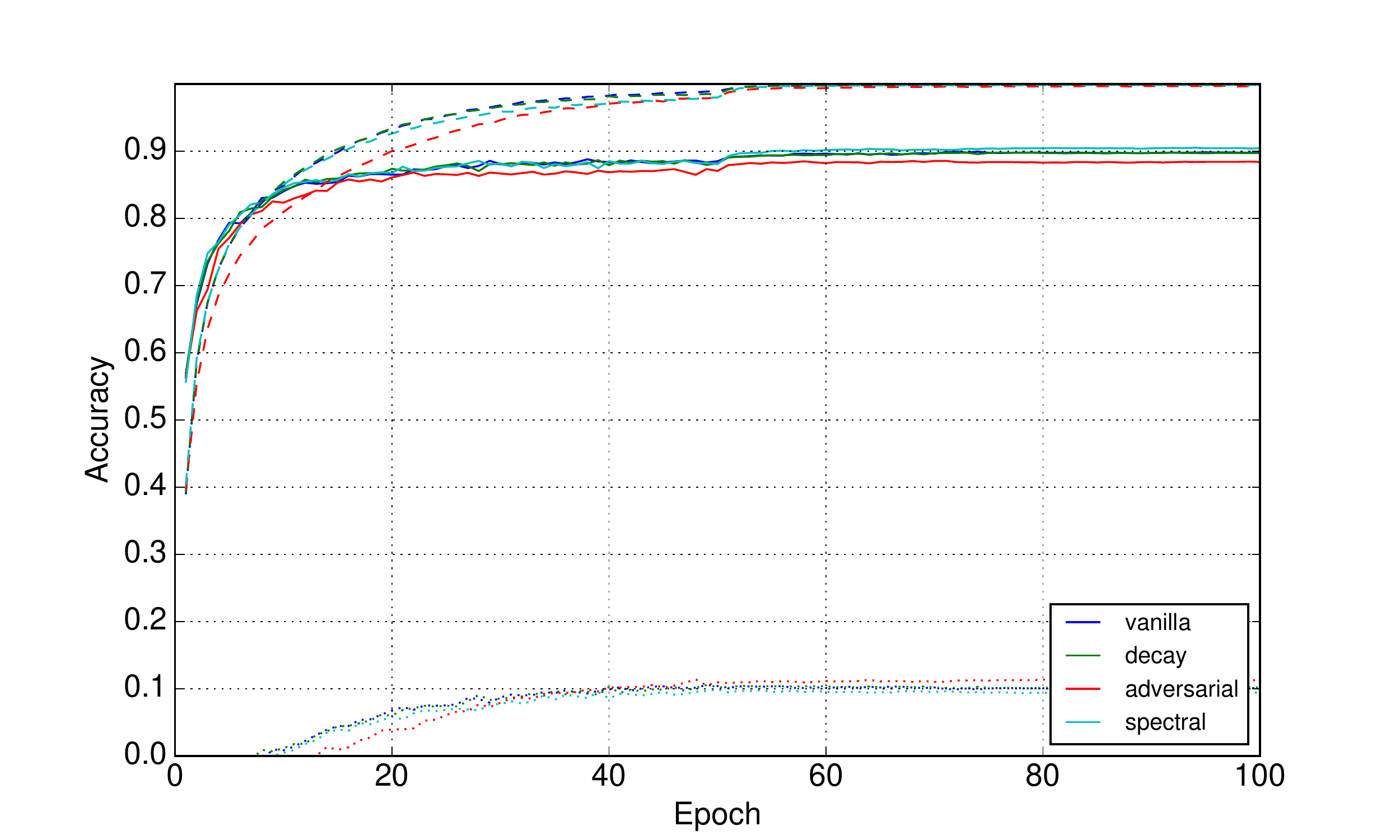}
  }
  \subfigure[VGGNet ($B=4096$)]{
    \includegraphics[width=.48\hsize]{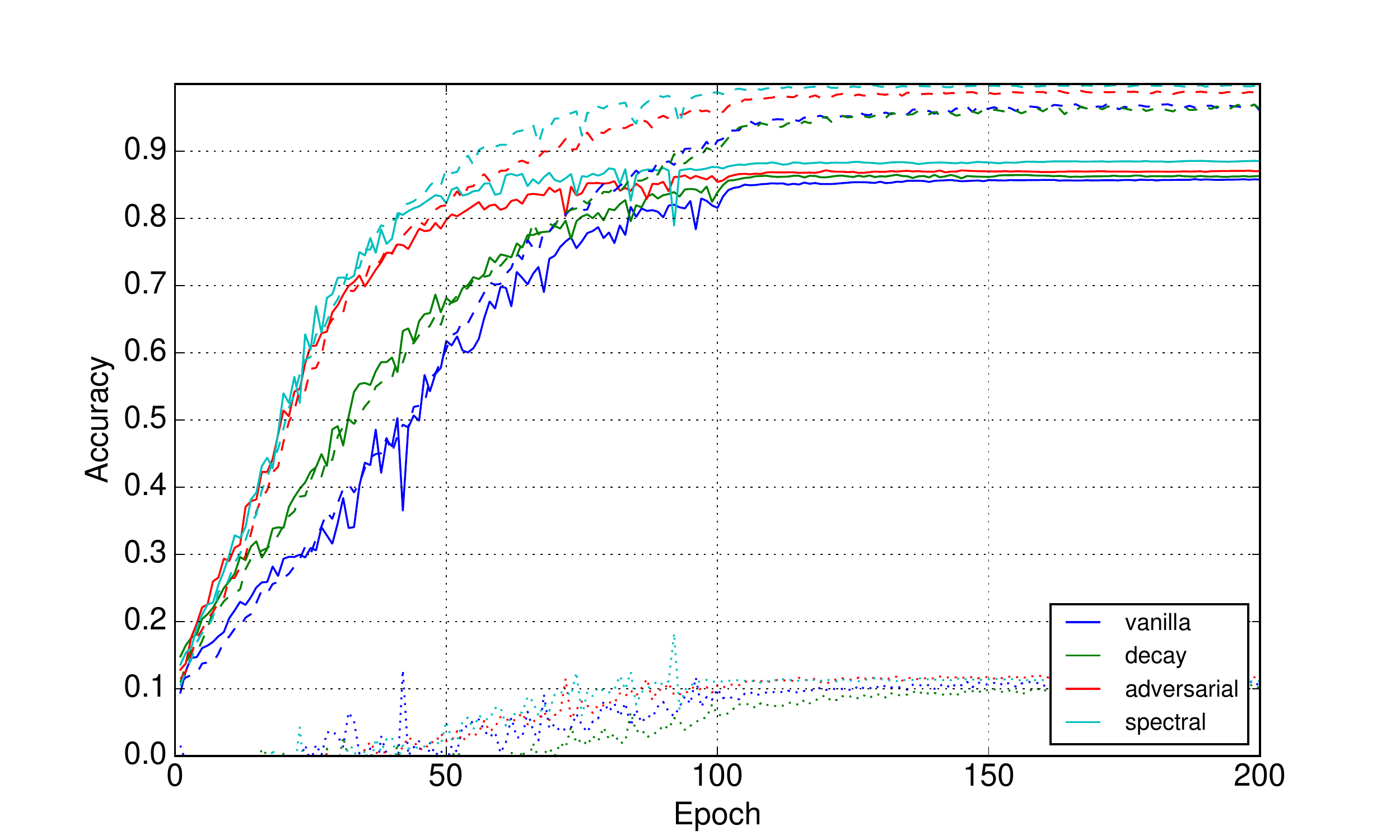}
  }
  \subfigure[NIN ($B=64$)]{
    \includegraphics[width=.48\hsize]{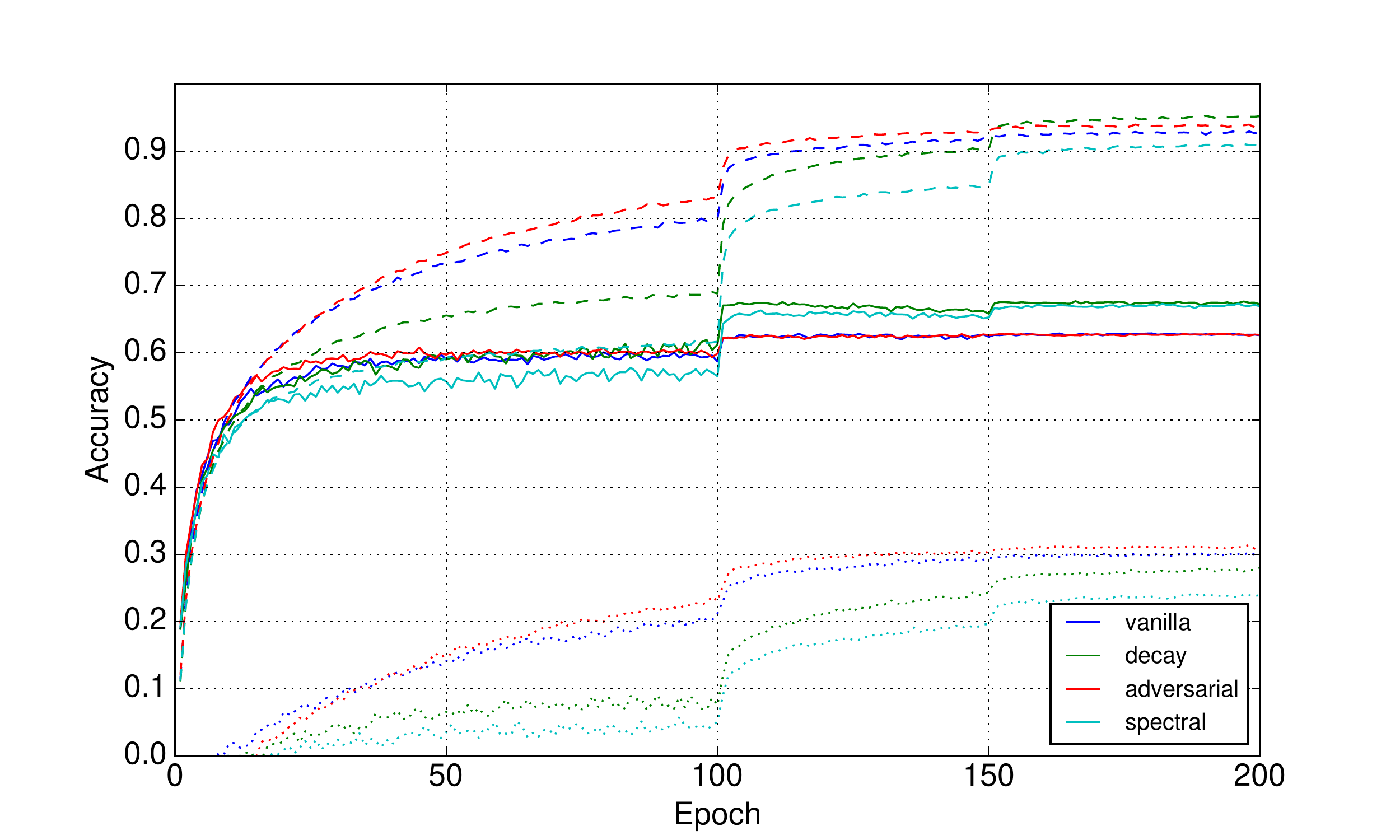}
  }
  \subfigure[NIN ($B=4096$)]{
    \includegraphics[width=.48\hsize]{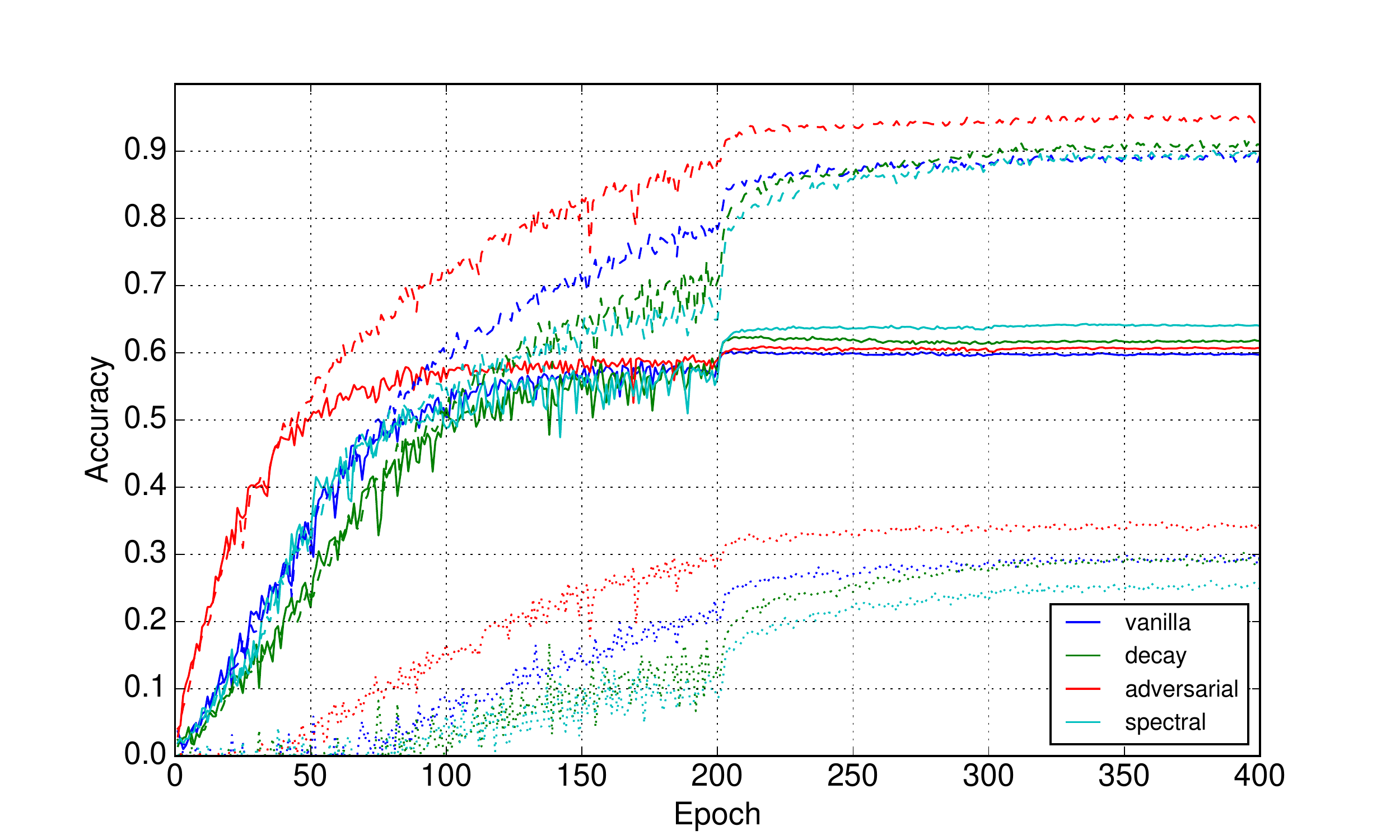}
  }
  \subfigure[DenseNet on STL-10 ($B=64$)]{
    \includegraphics[width=.48\hsize]{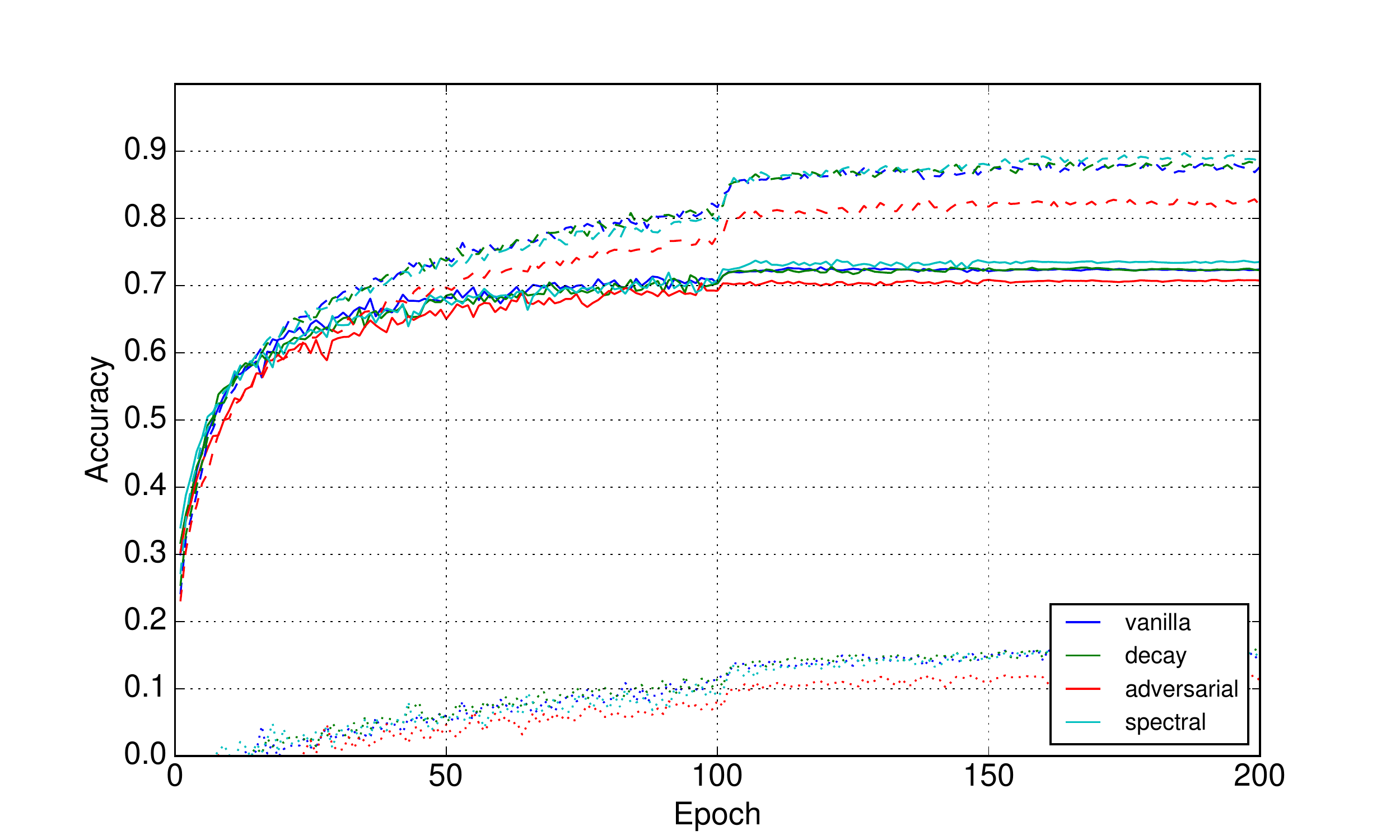}
  }
  \subfigure[DenseNet on STL-10 ($B=4096$)]{
    \includegraphics[width=.48\hsize]{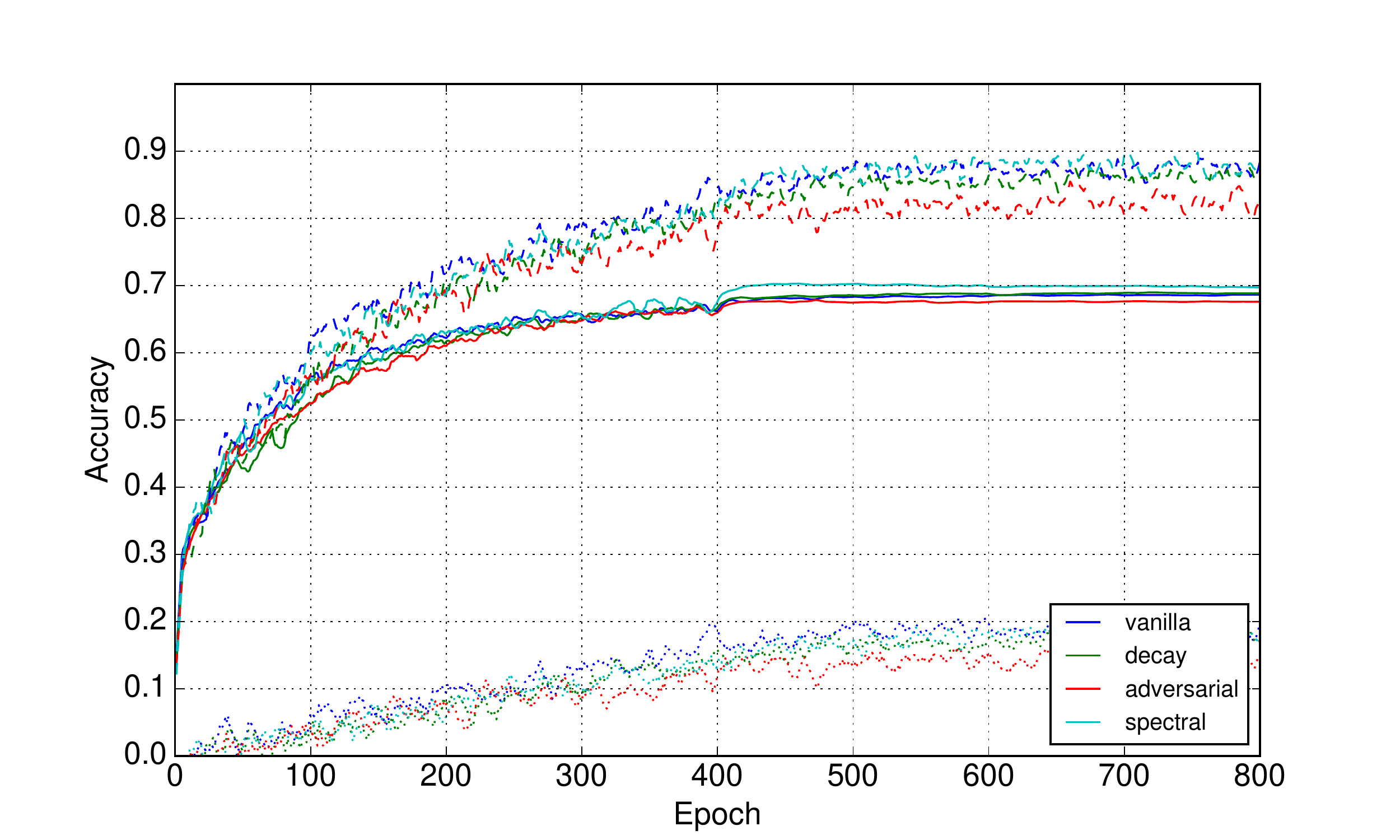}
  }
  \caption{Accuracy. Solid, dashed, and dotted lines indicate testing accuracy, training accuracy, and the generalization gap, respectively.}\label{fig:accuracy-appendix}
\end{figure*}

The training curves for the VGGNet, NIN, and DenseNet models on the STL-10 dataset are shown in Figure~\ref{fig:accuracy-appendix}.
We can observe that, in every setting, \textsf{spectral} shows the smallest generalization gap or the best test accuracy, which demonstrates that \textsf{spectral} can effectively reduce the generalization gap without suppressing the model complexity significantly.



\section{Sensitivity to the perturbation of the input}\label{apx:sensitivity}

\begin{figure*}
  \centering
  \subfigure[VGGNet (Train)]{
    \includegraphics[width=.475\hsize]{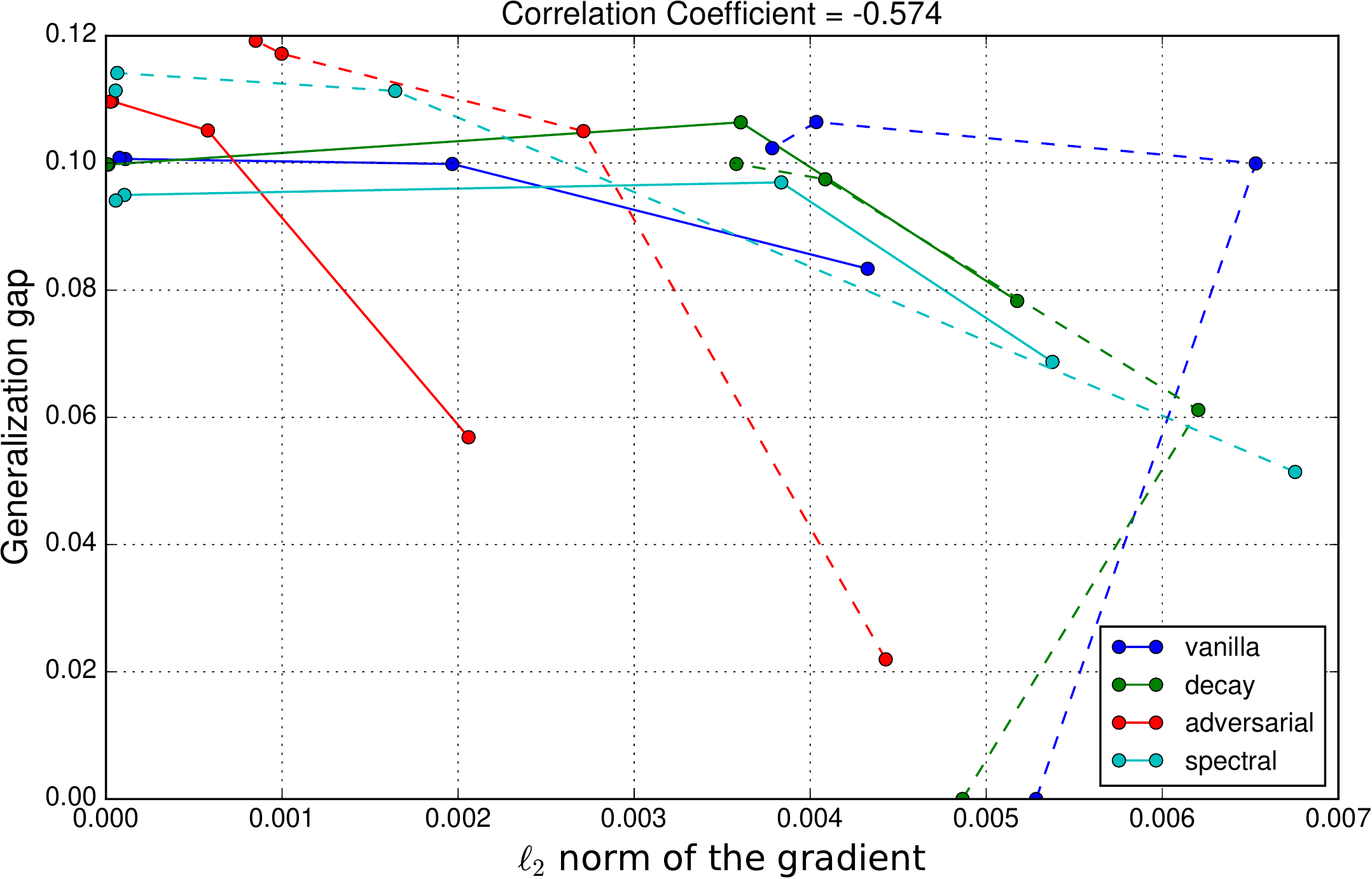}
  }
  \subfigure[VGGNet (Test)]{
    \includegraphics[width=.475\hsize]{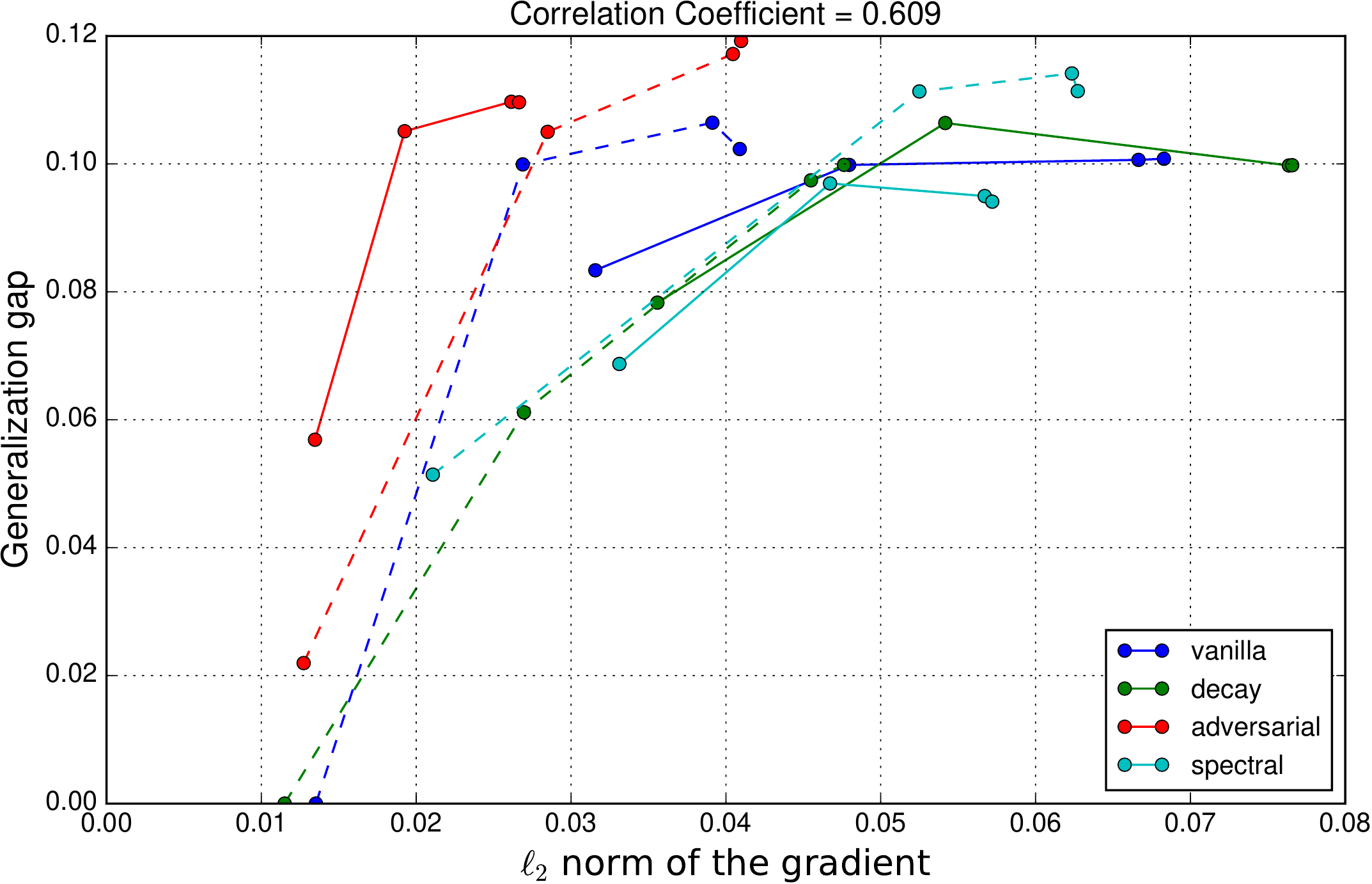}
  }

  \subfigure[NIN (Train)]{
    \includegraphics[width=.475\hsize]{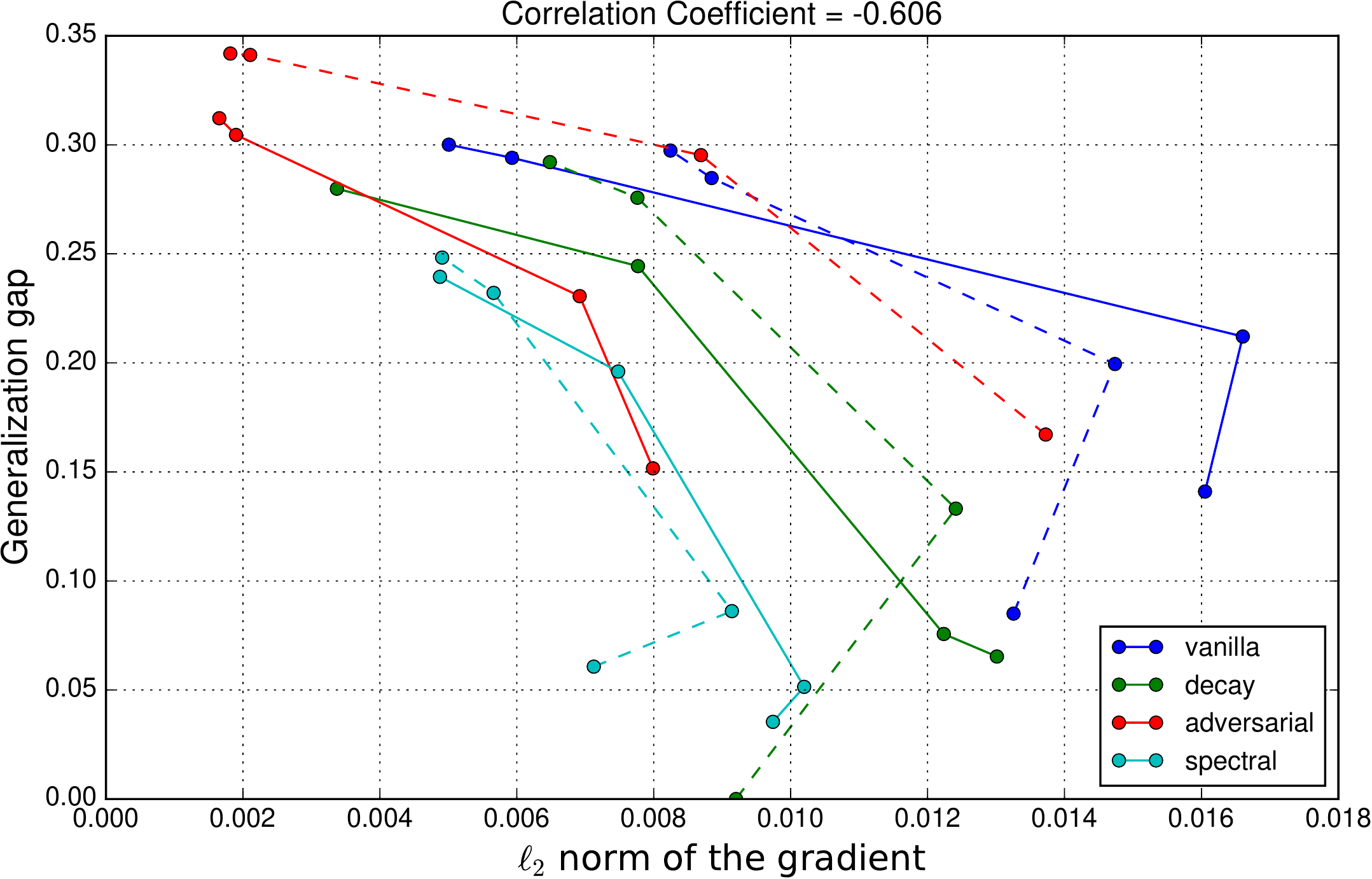}
  }
  \subfigure[NIN (Test)]{
    \includegraphics[width=.475\hsize]{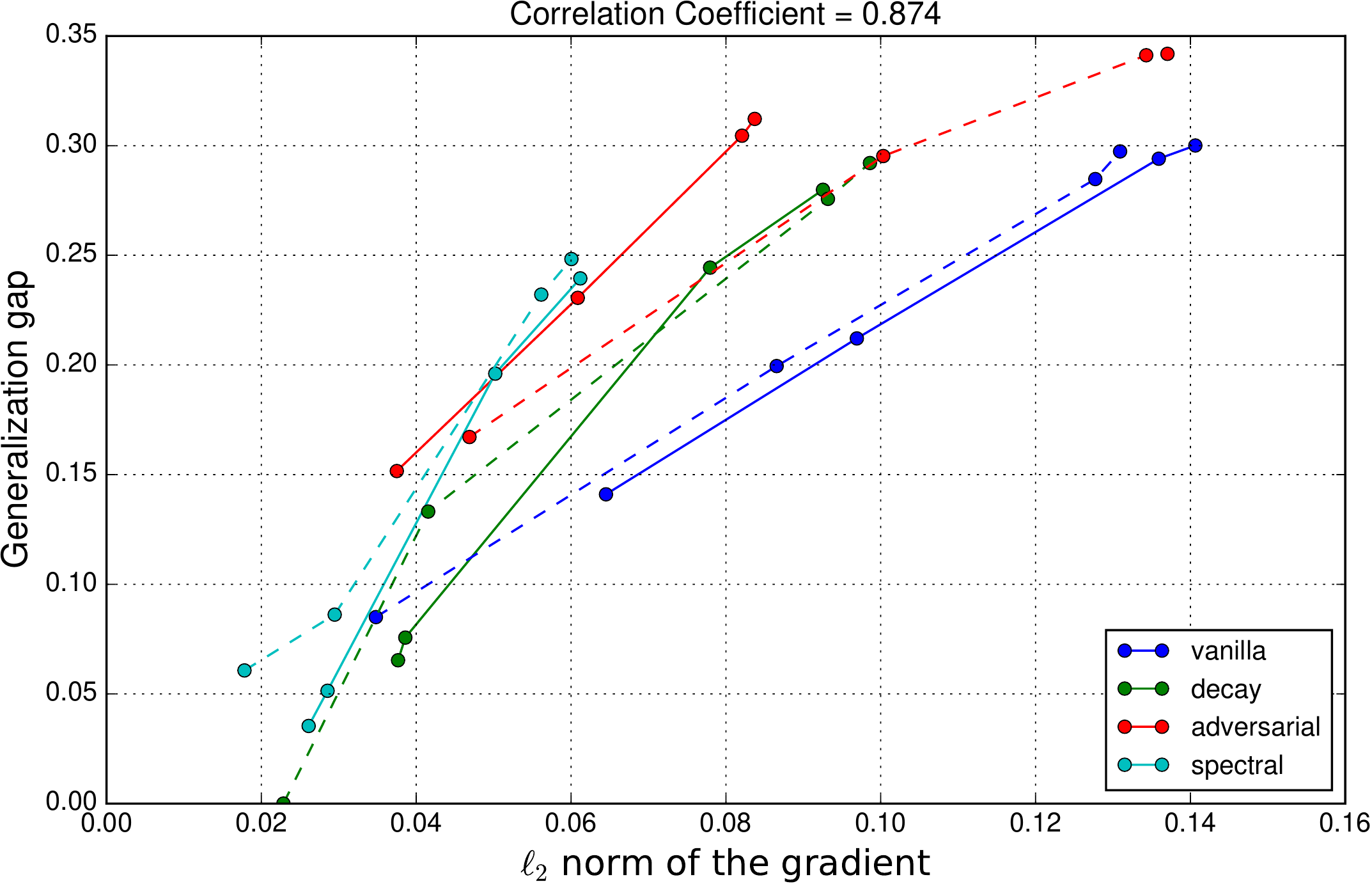}
  }

  \subfigure[DenseNet on STL-10 (Train)]{
    \includegraphics[width=.475\hsize]{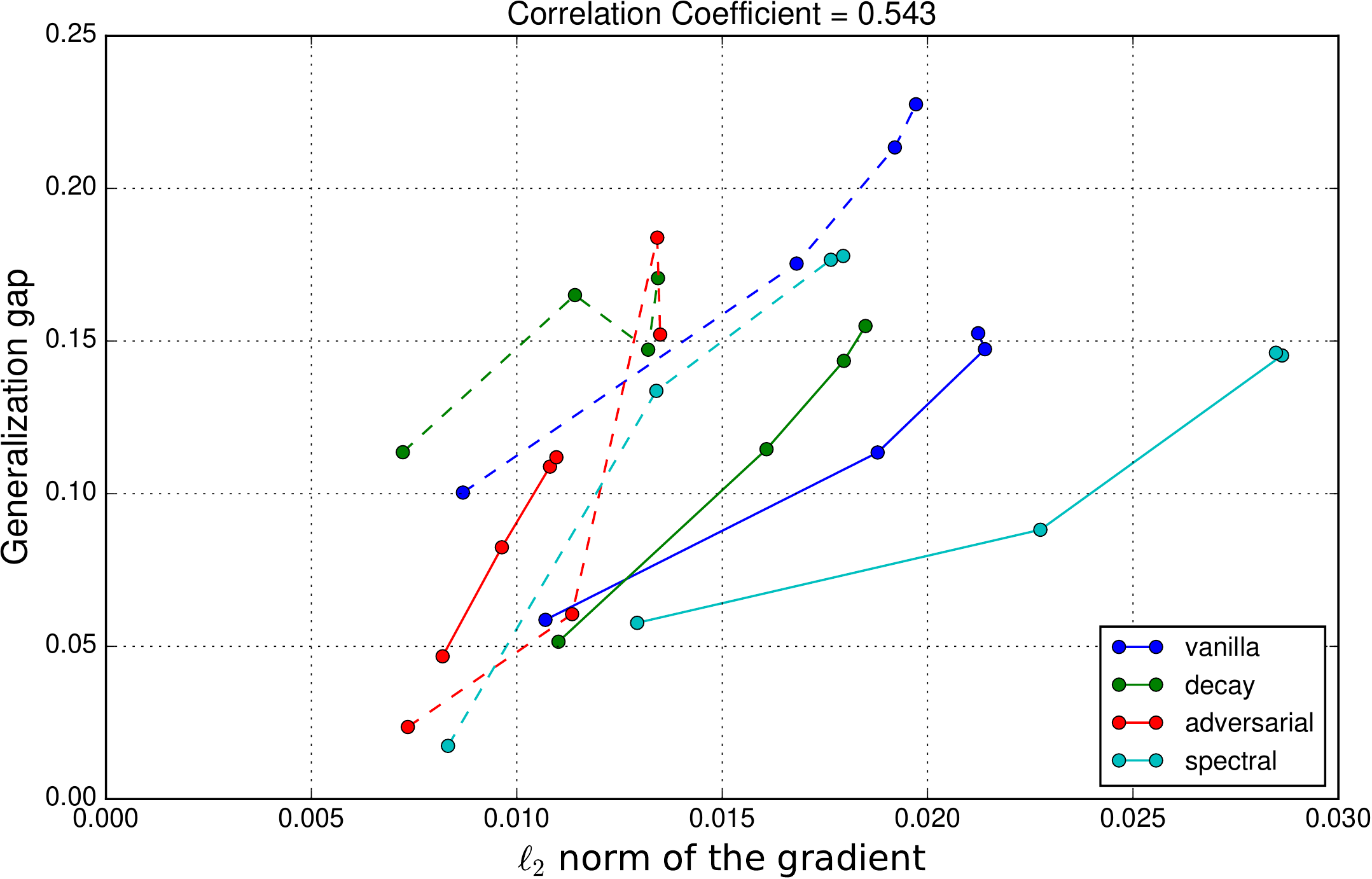}
  }
  \subfigure[DenseNet on STL-10 (Test)]{
    \includegraphics[width=.475\hsize]{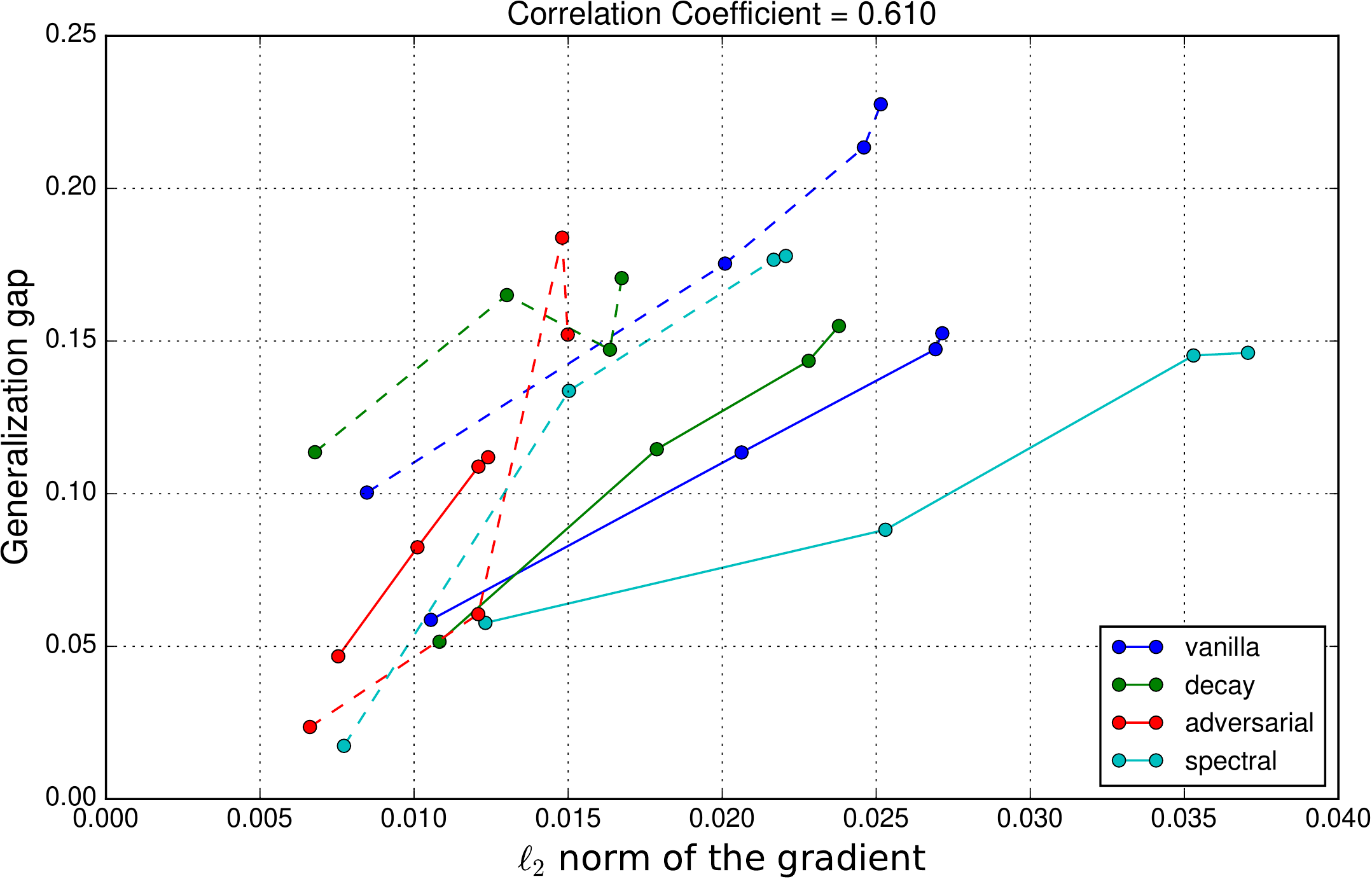}
  }

  \caption{Relation between the generalization gap and the $\ell_2$-norm of the gradient. The solid and dashed lines indicate the results for the small-batch and large-batch regimes, respectively.}\label{fig:grad-transition-appendix}
\end{figure*}

Figure~\ref{fig:grad-transition-appendix} shows the transition of the $\ell_2$-norm of the gradient of the loss function, defined with the training data or test data, with respect to the input.
For every setting, the $\ell_2$-norm of the gradient with respect to the test data is well-correlated with the generalization gap.
On the contrary, for the VGGNet and NIN models, the $\ell_2$-norm of the gradient with respect to the training data does not predict generalization gap well.